\pdfoutput=1

\IfFileExists{./prepreamble-amspreprint.sty}{\RequirePackage[packages,theorems,changes]{./prepreamble-amspreprint}}{}

\makeatletter
\IfFileExists{./scoop-latex/scoop-packages.sty}{\providecommand*{\input@path}{}\edef\input@path{{./scoop-latex/}\input@path}}{}
\@namedef{ver@minted.sty}{}
\@namedef{opt@minted.sty}{}
\makeatother
\PassOptionsToPackage{style=authoryear-comp,backend=biber}{biblatex}
\PassOptionsToPackage{commentmarkup = footnote}{changes}    

\documentclass[english]{amsart}

\RequirePackage{biblatex}
\RequirePackage{ltxcmds}
\IfFileExists{./preamble-amspreprint.sty}{\RequirePackage[packages,theorems,changes]{./preamble-amspreprint}}{}

\usepackage{scoop-local}

\IfFileExists{./postpreamble-amspreprint.sty}{\RequirePackage[packages,theorems,changes]{./postpreamble-amspreprint}}{}

\makeatletter
\@ifpackageloaded{changes}{
\definechangesauthor[name = {Evelyn Herberg}, color = {red!80!black}]{EH}
\definechangesauthor[name = {Roland Herzog}, color = {blue!80!black}]{RH}
\definechangesauthor[name = {Frederik Köhne}, color = {orange!80!black}]{FK}
}{}
\makeatother        

\makeatletter
\@ifpackageloaded{hyperref}{%
	\hypersetup{
		pdftitle = {Time Regularization in Optimal Time Variable Learning},
		pdfauthor = {Evelyn Herberg, Roland Herzog, Frederik Köhne},
		pdfkeywords = {deep learning, deep neural networks, network architecture, PyTorch},
		pdfcreator = {Created using the Scoop Template Engine version 1.2.0.}
	}
}{
	\pdfinfo{
		/Title (Time Regularization in Optimal Time Variable Learning)
		/Author (Evelyn Herberg, Roland Herzog, Frederik Köhne)
		/Subject ()
		/Keywords (deep learning, deep neural networks, network architecture, PyTorch)
		/Creator (Created using the Scoop Template Engine version 1.2.0.)
	}
}
\makeatother

\title[Time Regularization in Optimal Time Variable Learning]{Time Regularization in Optimal Time Variable Learning}

\author{Evelyn Herberg\orcidlink{0000-0003-2515-4818}}
\address[E. Herberg]{Interdisciplinary Center for Scientific Computing, Heidelberg University, 69120 Heidelberg, Germany}
\email{evelyn.herberg@iwr.uni-heidelberg.de}
\urladdr{https://scoop.iwr.uni-heidelberg.de}

\author{Roland Herzog\orcidlink{0000-0003-2164-6575}}
\address[R. Herzog]{Interdisciplinary Center for Scientific Computing, Heidelberg University, 69120 Heidelberg, Germany}
\email{roland.herzog@iwr.uni-heidelberg.de}
\urladdr{https://scoop.iwr.uni-heidelberg.de}

\author{Frederik Köhne\orcidlink{0009-0008-6185-9675}}
\address[F. Köhne]{Department of Mathematics, University of Bayreuth, 95440 Bayreuth, Germany}
\email{frederik.koehne@uni-bayreuth.de}
\urladdr{https://num.math.uni-bayreuth.de/en/team/frederik-koehne/}

\date{\today}

\dedicatory{}

\begin{document}

\begin{abstract}
Recently, optimal time variable learning in deep neural networks (DNNs) was introduced in \cite{AntilDiazHerberg:2022:1}. 
In this manuscript we extend the concept by introducing a regularization term that directly relates to the time horizon in discrete dynamical systems. 
Furthermore, we propose an adaptive pruning approach for Residual Neural Networks (ResNets), which reduces network complexity without compromising expressiveness, while simultaneously decreasing training time.   
The results are illustrated by applying the proposed concepts to classification tasks on the well known MNIST and Fashion MNIST data sets. 
Our PyTorch code is available on \url{https://github.com/frederikkoehne/time_variable_learning}, \cite{Koehne:2023:1}.

\end{abstract}

\keywords{deep learning, deep neural networks, network architecture, PyTorch}

\makeatletter
\ltx@ifpackageloaded{hyperref}{%
\subjclass[2010]{}
}{%
\subjclass[2010]{}
}
\makeatother

\maketitle

\section{Introduction}
\label{section:introduction}

Some deep neural networks (DNNs) can be understood as time discretizations of non-linear ordinary differential equations (ODEs).
For instance, a simplified version of ResNet \cite{HeZhangRenSun:2016:3} corresponds to an explicit Euler time stepping, and Fractional DNNs \cite{AntilKhatriLoehnerVerma:2020:1,AntilElmanOnwuntaVerma:2021:1} correspond to a discretization of the Caputo fractional time derivatives with a finite difference approach \cite{LinXu:2007:1,LinLiXu:2011:1}. 
For the ResNet architecture, this idea was introduced in \cite{HaberRuthottoHolthamJun:2017:1,HaberRuthotto:2017:2,RuthottoHaber:2018:1}.
Furthermore, there exist several other works which consider learning with DNNs as discretizations of optimization problems constrained by dynamical systems, \eg, \cite{E:2017:1,ChenRubanovaBettencourtDuvenaud:2018:1,LuZhongLiDong:2018:1,BenningCelledoniEhrhardtOwrenSchoenlieb:2019:1,LiuMarkowich:2020:1}, and many more which are not mentioned here. 
One of the earlier works in this direction is \cite{Cessac:2010:1}.

As in \cite{AntilDiazHerberg:2022:1}, we introduce additional variables $\tau{\ell}$, which can be interpreted as time step sizes; see \eqref{eq:resnet} below. 
When a time discretization of a non-linear ODE is employed, often a time interval with a fixed time horizon $T$ is considered. 
We take this idea to the DNN setting by enforcing that the variable time step sizes sum up to~$T$. 

The plan of this manuscript is as follows. 
We first introduce the learning problem and the network architectures under consideration. 
Subsequently, regularization concepts are discussed and the adaptive pruning approach is explained. 
Finally, we present numerical results for the MNIST and Fashion MNIST data sets.

\section{Problem Formulation}
\label{section:problem-formulation}

This section is a short recapitulation of the problem formulation in \cite{AntilDiazHerberg:2022:1}. 
Consider supervised deep learning, \ie, the problem
\begin{align}
	\text{Minimize}
	\quad
	&
	L \paren[big](){%
		\paren[big]\{\}{%
			\layeroutput{L}{i}, \datalabel{i}
		}_{i=1}^N; \theta 
	} 
	\notag
	\\
	\text{subject to}
	\quad
	&
	\layeroutput{L}{i} 
	= 
	\cF(\datafeature{i}; \theta) 
	\quad 
	\text{for all }
	i = 1, \ldots, N
	.
	\label{eq:problemP}\tag{P}
\end{align}
The variable $\theta$ subsumes all learning (optimization) variables.
Here, $\paren[big]\{\}{(\datafeature{i},\datalabel{i})}_{i=1}^N$ is the given data, consisting of inputs (features) and outputs (labels). 
The function~$\cF$ represents a deep neural network (DNN), to be specified, with output~$\layeroutput{L}$ on layer~$L$. 
The goal is to match the outputs $\layeroutput{L}{i}$ with the labels~$\datalabel{i}$ for all data points~$i$, which will be modeled through the loss function $L$. 
For multiclass classification tasks, we consider the negative log-likelihood as loss function and insert the defining constraint for $\layeroutput{L}{i}$, \ie
\begin{equation*}
	L(\theta)
	\coloneqq
	\frac{1}{N} \sum_{i=1}^N -\log \paren[auto](){%
		P \paren[big](){%
			\layeroutput{L}{i} = \datalabel{i} | \datafeature{i}, \theta
		}
	}
	.
\end{equation*} 
Let us remark that by employing a Bernoulli distribution and an implicit softmax activation function in the last layer, we can replace the negative log-likelihood loss function with a cross entropy loss function in our implementations. 
This is a standard approach in multiclass classification. 
For more details, we refer the reader to, \eg, \cite[Section~2.4]{Herberg:2023:1}.

The network propagation function~$\cF$ represents a network of depth~$L$, which has an input layer, $L-1$ hidden layers and an output layer.
Let the input coincide with the given data~$\datafeature \eqqcolon \layeroutput{0} \in \R^{n_0}$, and define feature vectors $\layeroutput{\ell} \in \R^{n_\ell}$ for layers~$\ell = 1, \ldots, L$. 
Every layer has an associated layer function $\layerfunction{\ell} \colon \R^{n_\ell} \rightarrow \R^{n_{\ell + 1}}$, 
so that we can write the DNN as a concatenation of layer functions
\begin{equation*}
	\cF \colon \R^{n_0} \rightarrow \R^{n_L}
	,
	\quad 
	\cF 
	\coloneqq
	\layerfunction{L-1} \circ \layerfunction{L-2} \circ \ldots \circ \layerfunction{0}
	.
\end{equation*}
For the layer functions we consider simplified ResNet 
\cite{HeZhangRenSun:2016:3}
with the time variable $\tau^{\ell}$ learning framework \cite[Section 5.1]{AntilDiazHerberg:2022:1}
\begin{equation}
	\label{eq:resnet}
	\layeroutput{\ell}
	=
	\layerfunction{\ell-1} (\layeroutput{\ell-1}) 
	{}\coloneqq{}
	\layeroutput{\ell-1} + \tau{\ell-1} \, \sigma 
	\paren[big](){%
		\weight{\ell-1} \layeroutput{\ell - 1} + \bias{\ell - 1} 
	},
\end{equation}
and Fractional DNN \cite{AntilKhatriLoehnerVerma:2020:1,AntilElmanOnwuntaVerma:2021:1} also with time variable learning framework \cite[Section 5.2]{AntilDiazHerberg:2022:1}
\makeatletter
\ltx@ifclassloaded{amsart}{%
\begin{multline}
	\label{eq:fractional-dnn}
	\layeroutput{\ell} 
	=
	\layerfunction{\ell-1} (\layeroutput{\ell-1}) 
	\\
	{}\coloneqq{}
	\layeroutput{\ell-1} - \sum_{j=0}^{\ell - 2} a_{\ell-1,j } 
	\paren[big](){%
		\layeroutput{j+1} - \layeroutput{j}
	}
	+ 
	(\tau{\ell-1})^{\gamma}\, \Gamma(2-\gamma) \, \sigma 
	\paren[auto](){%
		\weight{\ell-1} \layeroutput{\ell - 1} + \bias{\ell - 1} 
	},
\end{multline}
}{%
\begin{equation}
	\label{eq:fractional-dnn}
	\layeroutput{\ell} 
	=
	\layerfunction{\ell-1} (\layeroutput{\ell-1}) 
	{}\coloneqq{}
	\layeroutput{\ell-1} - \sum_{j=0}^{\ell - 2} a_{\ell-1,j } 
	\paren[big](){%
		\layeroutput{j+1} - \layeroutput{j}
	}
	+ 
	(\tau{\ell-1})^{\gamma}\, \Gamma(2-\gamma) \, \sigma 
	\paren[auto](){%
		\weight{\ell-1} \layeroutput{\ell - 1} + \bias{\ell - 1} 
	},
\end{equation}
}
\makeatother
where $\Gamma(\cdot)$ denotes Euler's Gamma function, and 
\begin{equation*}
	a_{\ell,j}
	\coloneqq
	\frac{(\tau{\ell})^\gamma}{\tau{j}} 
	\paren[Bigg](){%
		\paren[Big](){%
			\sum_{i=j}^\ell \tau{i}
		}^{1-\gamma} - 
		\paren[Big](){%
			\sum_{i=j+1}^{\ell} \tau{i} 
		}^{1-\gamma} 
	}
	.
\end{equation*}
Consequently, the learning variables~$\theta$ in problem \eqref{eq:problemP} comprise weights~$\paren[auto]\{\}{ \weight{\ell}}_{\ell=0}^{L-1}$, biases~$\paren[auto]\{\}{ \bias{\ell}}_{\ell=0}^{L-2}$ and the time step sizes~$\paren[auto]\{\}{ \tau{\ell}}_{\ell=0}^{L-2}$.

Due to the skip connections, the layer functions introduced above can only be applied when the widths of subsequent layers coincide. 
We therefore assume $n_1 = \ldots = n_{L-1}$ and make the following choices for $\layerfunction{0}$ and $\layerfunction{L-1}$, which will allow the network to work with arbitrarily sized data, \ie
\begin{align*}
	\layerfunction{0} (\layeroutput{0}) 
	&
	= 
	\begin{cases}
		\tau{0} \, \sigma 
		\paren[auto](){%
			\weight{0} \layeroutput{0} + \bias{0} 
		}
		&
		\text{in case of ResNet}
		,
		\\
		(\tau{0})^{\gamma}\, \Gamma(2-\gamma) \, \sigma 
		\paren[auto](){%
			\weight{0} \layeroutput{0} + \bias{0} 
		}
		&
		\text{in case of Fractional DNN}
		,
	\end{cases}
	\\ 
	\layerfunction{L-1} (\layeroutput{L-1}) 
	&
	= 
	\weight{L-1} \layeroutput{L-1}
	.
\end{align*}
Furthermore, in Fractional DNNs, we will account for the dimensional mismatch by starting the sum in \eqref{eq:fractional-dnn} at $j = 1$.
Alternatively, it would be possible to insert projection operators $P^{\ell}_0$ for $\ell = 1, \ldots, L-1$, so that $P^{\ell}_0 \layeroutput{0} \in \R^{n_\ell}$ for $\ell = 1, \ldots, L-1$.

\section{Time Regularization}
\label{section:time-regularization}

In the spirit of differential equations we introduce a time horizon~$T$ and the additional constraint
\begin{equation*}
	\sum_{\ell=0}^{L-2} \tau{\ell} 
	= 
	T
\end{equation*}
to our learning problem. 
Since we employ unconstrained optimization algorithms, we propose two options to incorporate the constraint. 
Either, we introduce a quadratic penalty term (here with a penalty factor of~$1$) in our loss function, \ie
\begin{equation*}
	L_T(\theta) 
	= 
	L(\theta) 
	+ 
	\frac{1}{2} \paren[Big](){T - \sum_{\ell=0}^{L-2} \tau{\ell}}^2
	,
\end{equation*}
or else we set $\tau{L-2} = T - \sum_{\ell=0}^{L-3} \tau{\ell}$. 
The latter approach often results in $\tau{L-2} <0$, which is not desirable when interpreting $\tau{L-2}$ as a time step size. 
Furthermore, for Fractional DNN in general, we have to ensure that $\tau{\ell} \neq 0$ holds for all $j$, in order to avoid degeneracy of $a_{\ell,j}$. 
Consequently, a projection step lends itself when working with Fractional DNN, that ensures $\tau{\ell} > 0$. 

Alternatively, we can also consider a standard regularization of the time stepsizes $\tau \coloneqq \paren[auto](){\tau{0}, \ldots, \tau{L-2}}^\top$. 
Here, we employ a sum norm, often referred to as $\ell^1$ regularization, \ie
\begin{equation*}
	L_1(\theta) 
	\coloneqq
	L(\theta) + \alpha \, \norm{\tau}_1
	,
\end{equation*}
where $\alpha \in \R$ is a penalty parameter.
This regularization term induces sparsity in $\tau$, which we can exploit with the following adaptive pruning technique.

\subsection*{Adaptive Pruning}

In problem formulations where $\cF$ represents a simplified ResNet architecture we observe an interesting property. 
If $\tau{\ell-1} = 0$ holds for some $\ell$, the definition \eqref{eq:resnet} of the layer function $\layerfunction{\ell}$ directly implies $\layeroutput{\ell} = \layeroutput{\ell-1}$.
This means that we have a redundant layer, which can be pruned without compromising the network's expressiveness. 
In our numerical examples we exploit this property and implement an adaptive pruning strategy; see \cref{fig:adaptive_pruning}. 
Whenever $\abs{\tau{\ell-1}} < \epsilon \sum_{j = 0}^{L-2}\abs{\tau{j}}$ is observed, the corresponding layer, including all of its variables, is deleted, and the training proceeds with the smaller network. 
We can incorporate this without noticeable computational cost. 
However, in the case that pruning is executed, we see an adaptive acceleration in our training due to the reduction in the number of variables. 
Additionally, adaptive pruning eliminates the search for the network depth hyperparameter~$L$, as long as we start training with a sufficiently deep network.

\section{Numerical Results}
\label{section:numerical-results}

Our implementation is available on 
\makeatletter
\ltx@ifclassloaded{amsart}{%
	\begin{center}
		\url{https://github.com/frederikkoehne/time_variable_learning} 
	\end{center}
	and can be cited as \cite{Koehne:2023:1}.%
}{%
\url{https://github.com/frederikkoehne/time_variable_learning} and can be cited as \cite{Koehne:2023:1}.%
}
\makeatother
We test our approach on the MNIST and Fashion MNIST data sets. 
Both data sets consist of $28 \times 28$ grayscale images that are sorted into 10~classes. 
For MNIST the classes are the digits from 0 to 9, while Fashion MNIST has clothing categories, \eg, class~2 is \enquote{pullover}. 
The data set provides \num{60000} data points for training and \num{10000} for testing.  
Consequently, we have input dimension $n_0 = 28^2 = 784$ and output dimension $n_L = 10$. 
We choose an architecture with 6~hidden layers, \ie, $L = 7$, and width $n_1 = \ldots = n_6 = 100$, in order to allow skip connections without the need for projection operators. 
This architecture has \num{130006} variables, 6~of which are the time step sizes $\tau{0}, \ldots, \tau{5}$. 
We employ the cross entropy loss and ReLU activation function, \ie, $\sigma(x) = \max\{0, x\}$, applied element-wise. 
The time horizon is set to $T = 1$ and we initialize $\tau^{\ell} = \frac{T}{L-1}$ for every $\ell$. 
In the experiments with fixed $\tau$, we set it to the same value for comparability. 
As optimization algorithm we employ mini-batch stochastic gradient descent with a mini-batch size of 100 and random sampling. 
Consequently, it takes 600~iterations to go over the whole training data set and complete an epoch. 
We run our experiments for 200~epochs, which amounts to $\num{1.2e5}$ iterations. 
The experiments have been conducted using an NVIDIA GeForce RTX~3070 GPU and the typical runtime per experiment is a few minutes. 

\begin{figure}[h!]
	\centering 
	\subfigure[MNIST]{\label{fig:a_1}\includegraphics[width=0.495\textwidth]{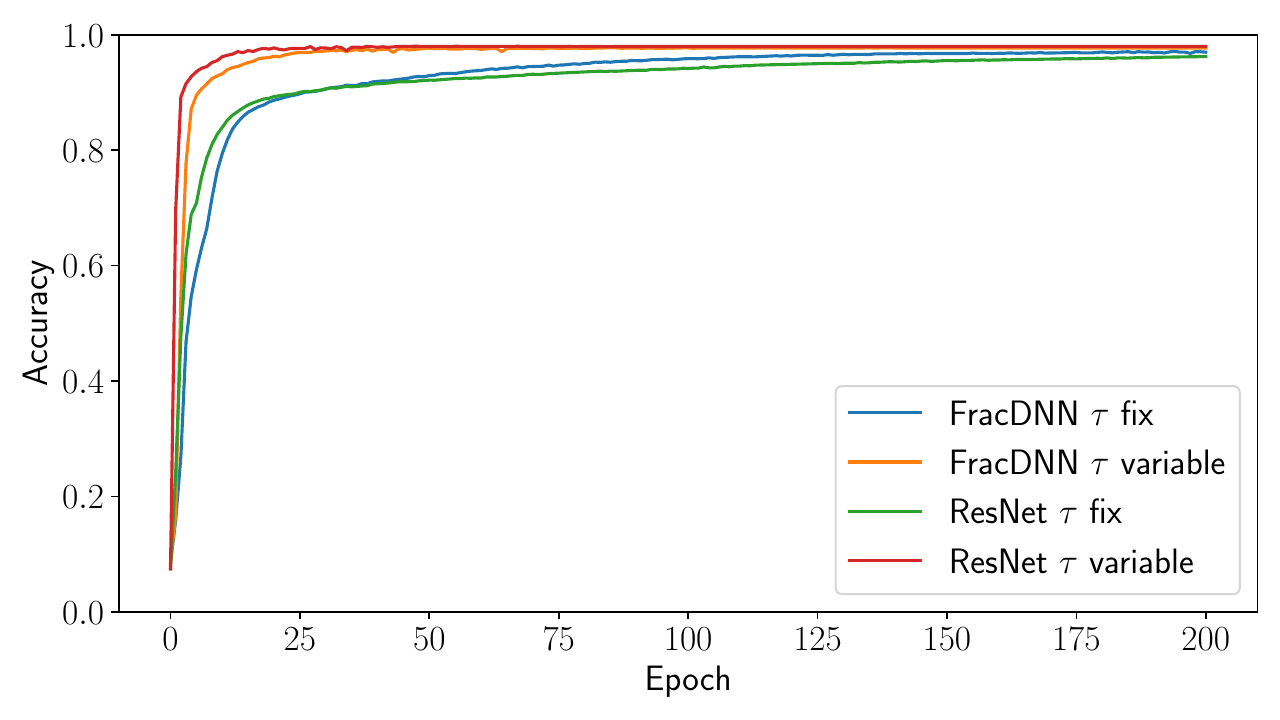}}
	\subfigure[Fashion MNIST]{\label{fig:a_2}\includegraphics[width=0.495\textwidth]{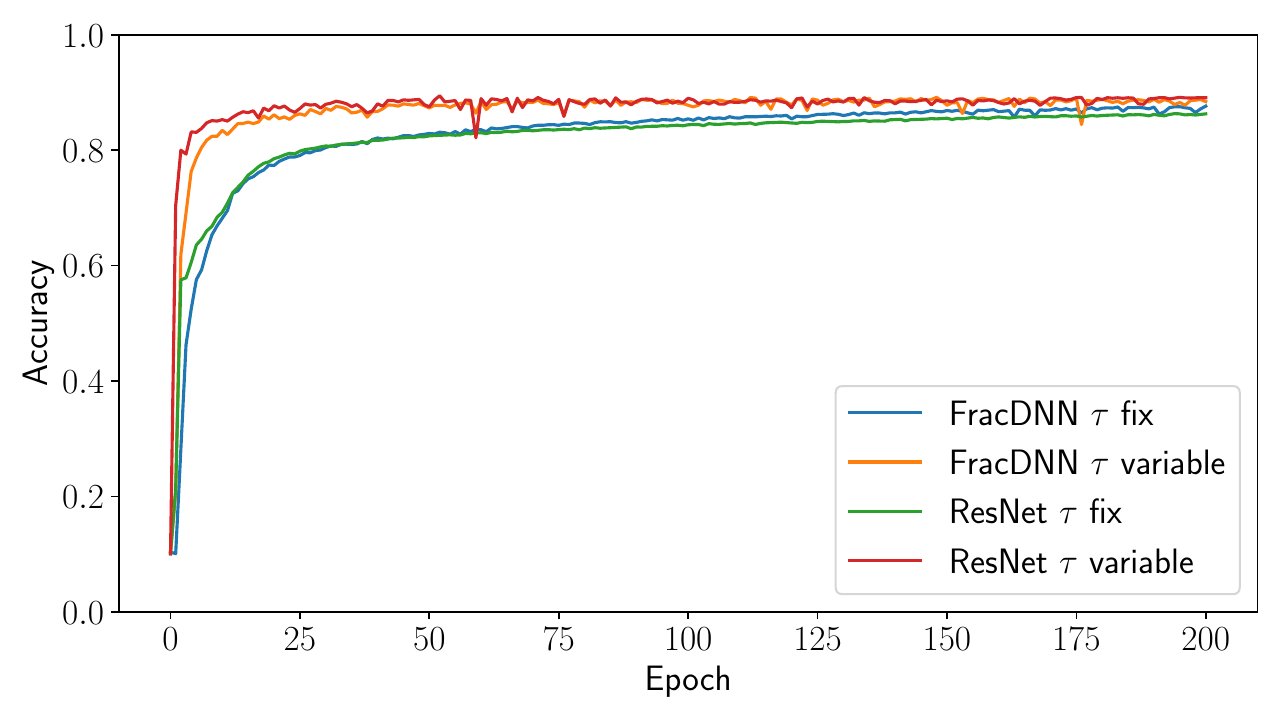}}
	\caption{Accuracies compared for trainable and fixed $\tau$ with different network architectures.}
	\label{fig:accuracies}
\end{figure}

In \cref{fig:accuracies} we see that ResNet and Fractional DNN with variable $\tau$ attain higher accuracies more quickly during training both for MNIST and Fashion MNIST data set, compared to the respective architectures with fixed time step size~$\tau$.
This coincides with the findings in \cite{AntilDiazHerberg:2022:1}, where training was also found to be faster by introducing time variable learning. 
This is remarkable since only 6~variables need to be added in comparison to the architecture with a fixed $\tau$.

\begin{figure}[ht]
	\centering 
	\subfigure[MNIST]{\label{fig:a_t_f_1}
		\includegraphics[width=0.495\textwidth]{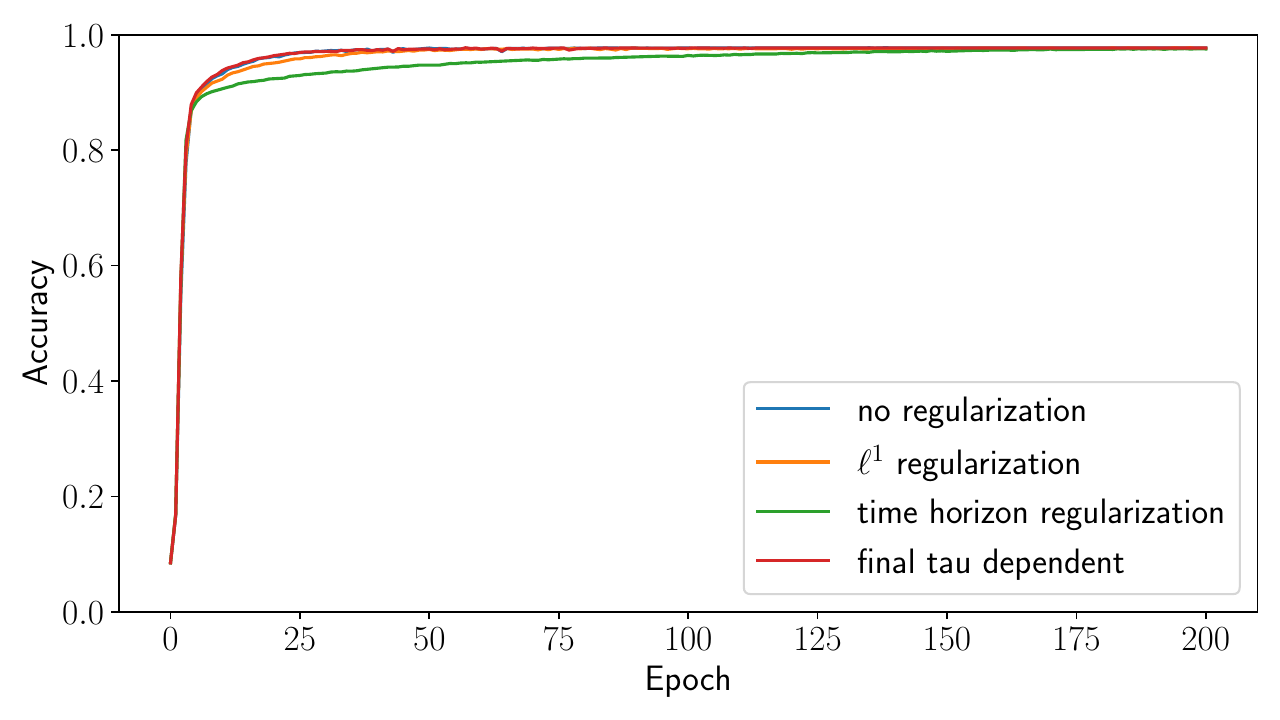}
		\includegraphics[width=0.495\textwidth]{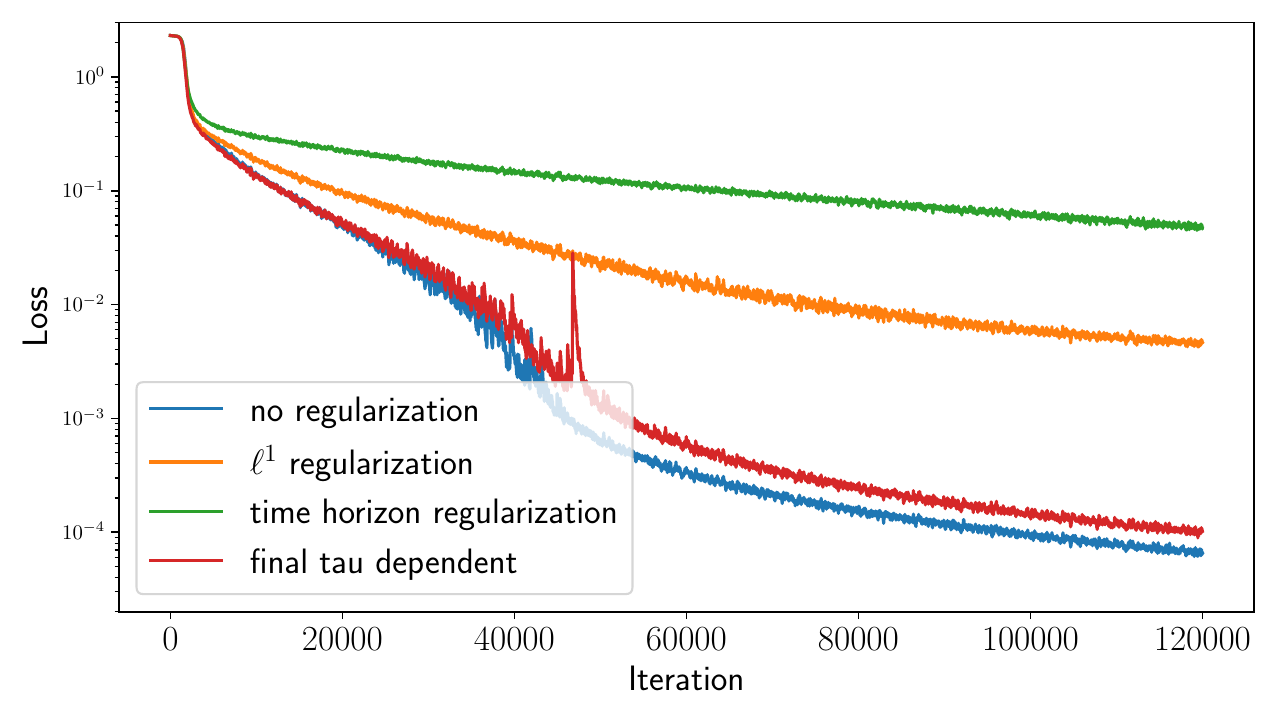}
	}
	\subfigure[Fashion
	MNIST]{\label{fig:a_t_f_3}
		\includegraphics[width=0.495\textwidth]{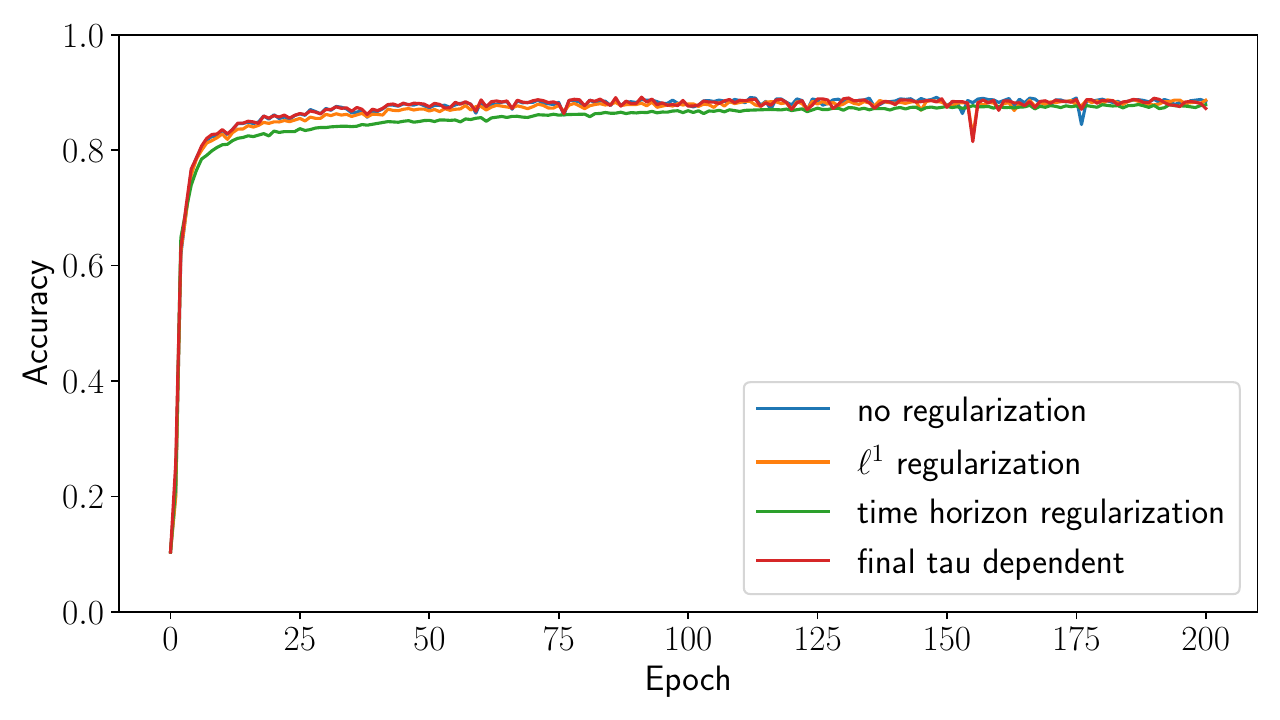}
		\includegraphics[width=0.495\textwidth]{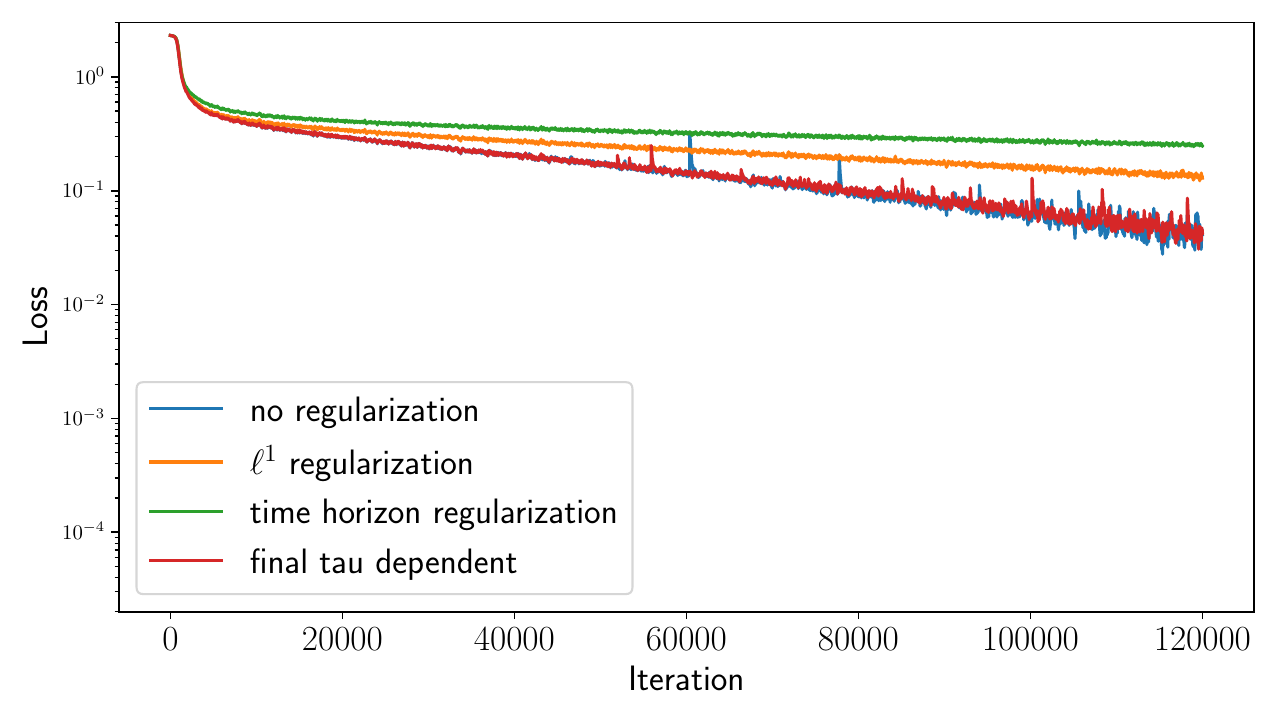}
	}
	\caption{Accuracies and cross entropy losses compared for different setups with trainable $\tau$ and Fractional DNN architecture.}
	\label{fig:accuracies_trainable_fdnn}
\end{figure}

\begin{figure}[ht]
	\centering 
	\subfigure[no regularization]{\label{fig:t_b_1}\includegraphics[width=0.32\textwidth]{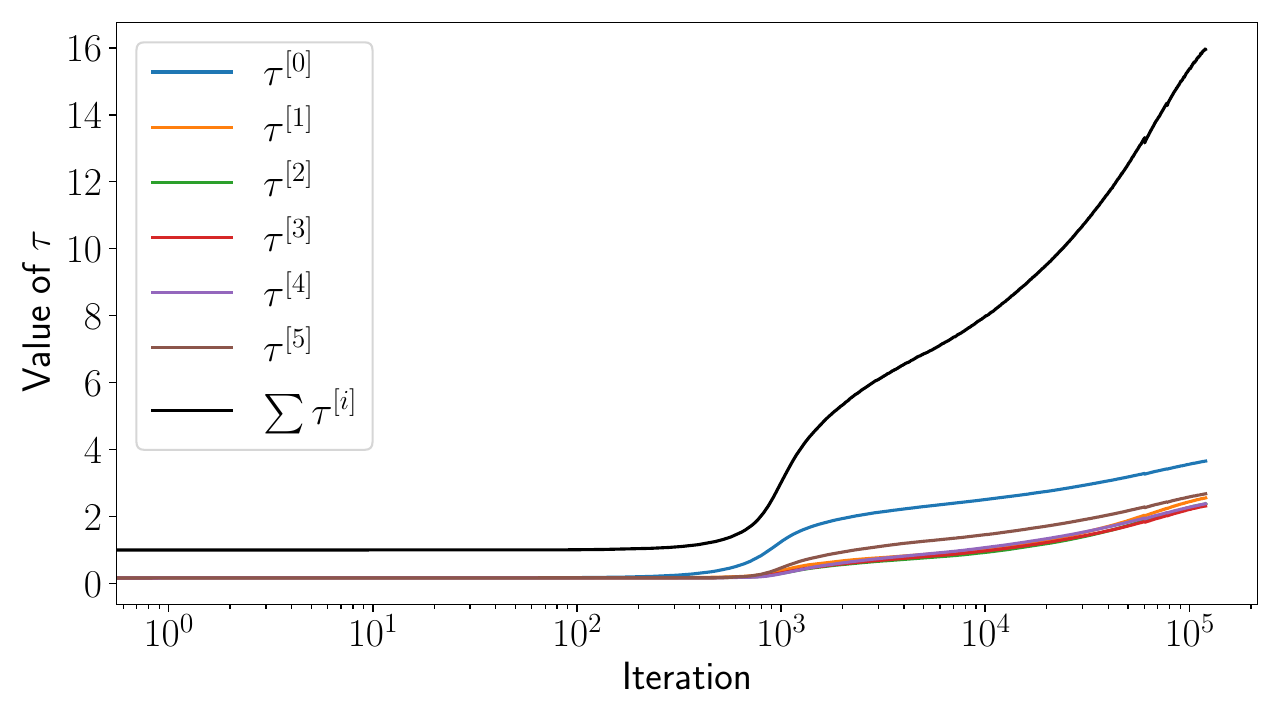}}
	\subfigure[$\ell^1$ regularization]{\label{fig:t_b_2}\includegraphics[width=0.32\textwidth]{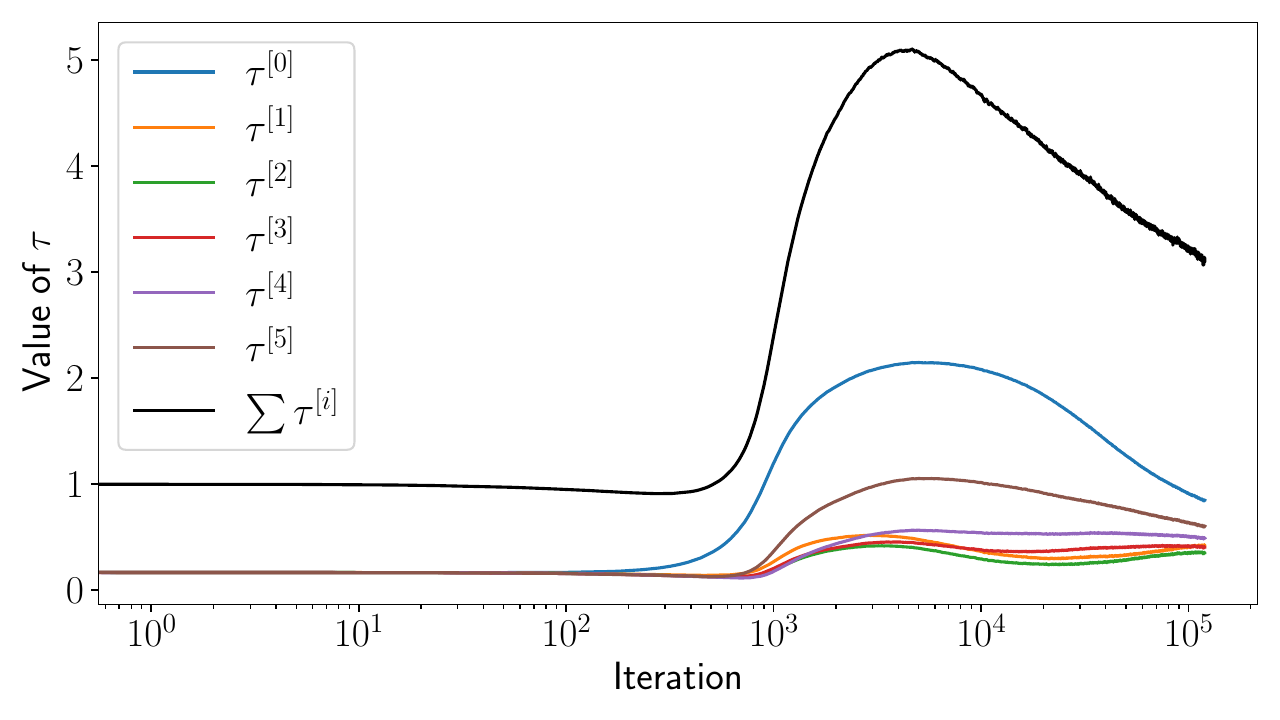}}
	\subfigure[time horizon regularization]{\label{fig:t_b_3}\includegraphics[width=0.32\textwidth]{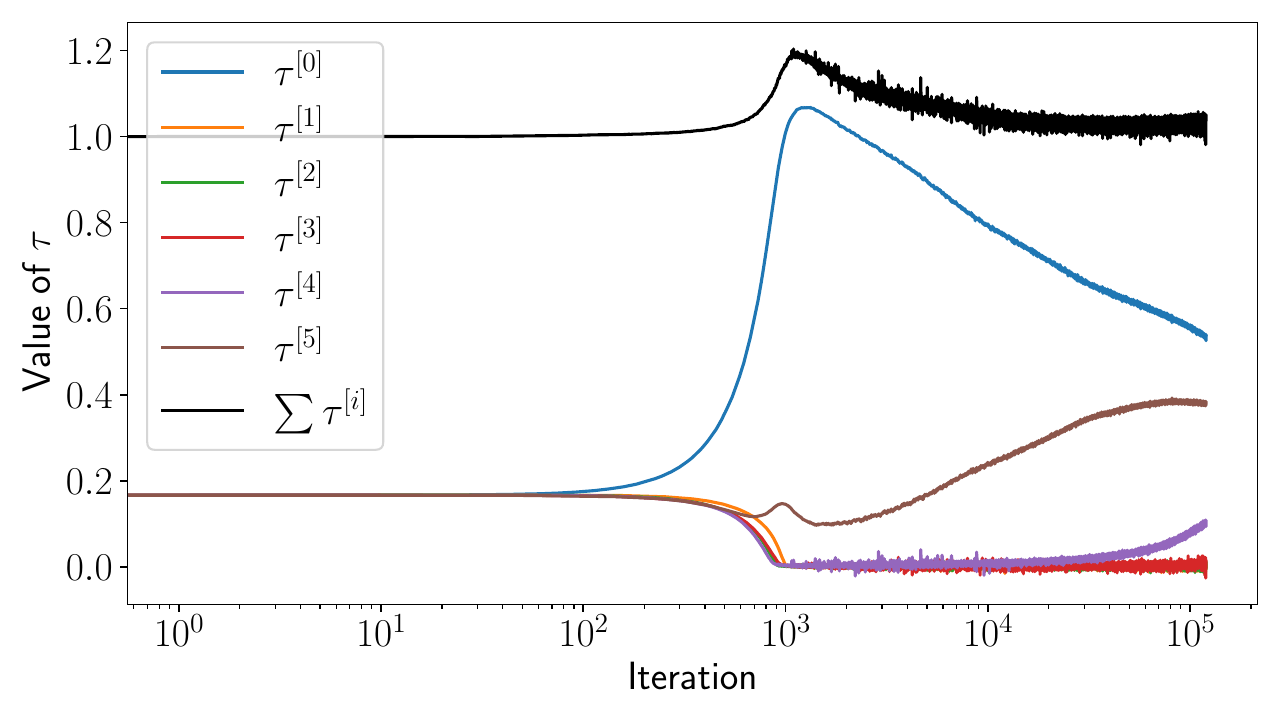}}
	\subfigure[final $\tau$ dependent]{\label{fig:t_b_4}\includegraphics[width=0.32\textwidth]{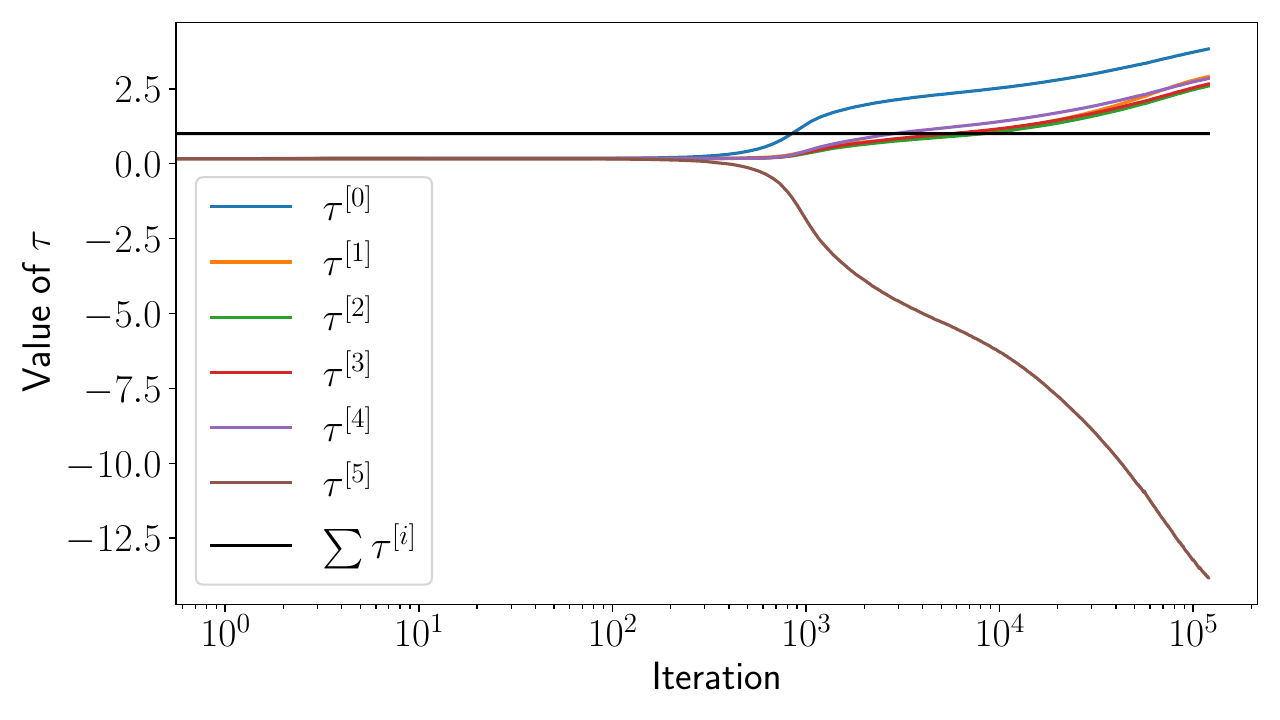}}
	\subfigure[$\ell^1$ and time horizon regularization]{\label{fig:t_b_5}\includegraphics[width=0.32\textwidth]{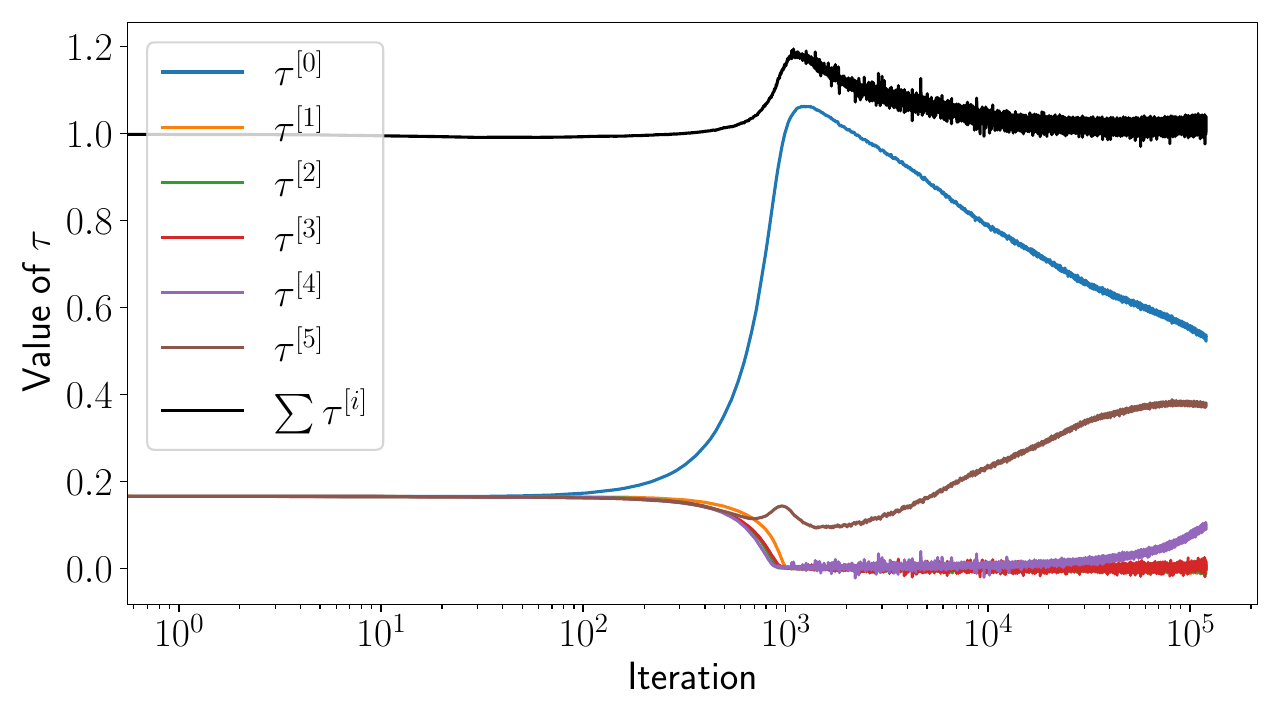}}
	\caption{Development of $\tau$ values over iterations for Fractional DNN architecture with different regularization terms while learning to classify Fashion MNIST data. Please note that the scale on the vertical axis differs from plot to plot.}
	\label{fig:tau_behavior}
\end{figure}

For the time variable learning we have compared no regularization with the different regularization approaches, \ie, $\ell^1$ and time horizon regularization and also architectures where the final $\tau{5}$ is computed from all previous time step sizes $\tau{0}, \ldots, \tau{4}$. 
For the upcoming experiments, we always choose $\alpha = 0.01$ when considering $\ell^1$ regularization.  
\begin{figure}[ht]
	\centering 
		\subfigure[MNIST]{\label{fig:a_t_r_1}
	\includegraphics[width=0.495\textwidth]{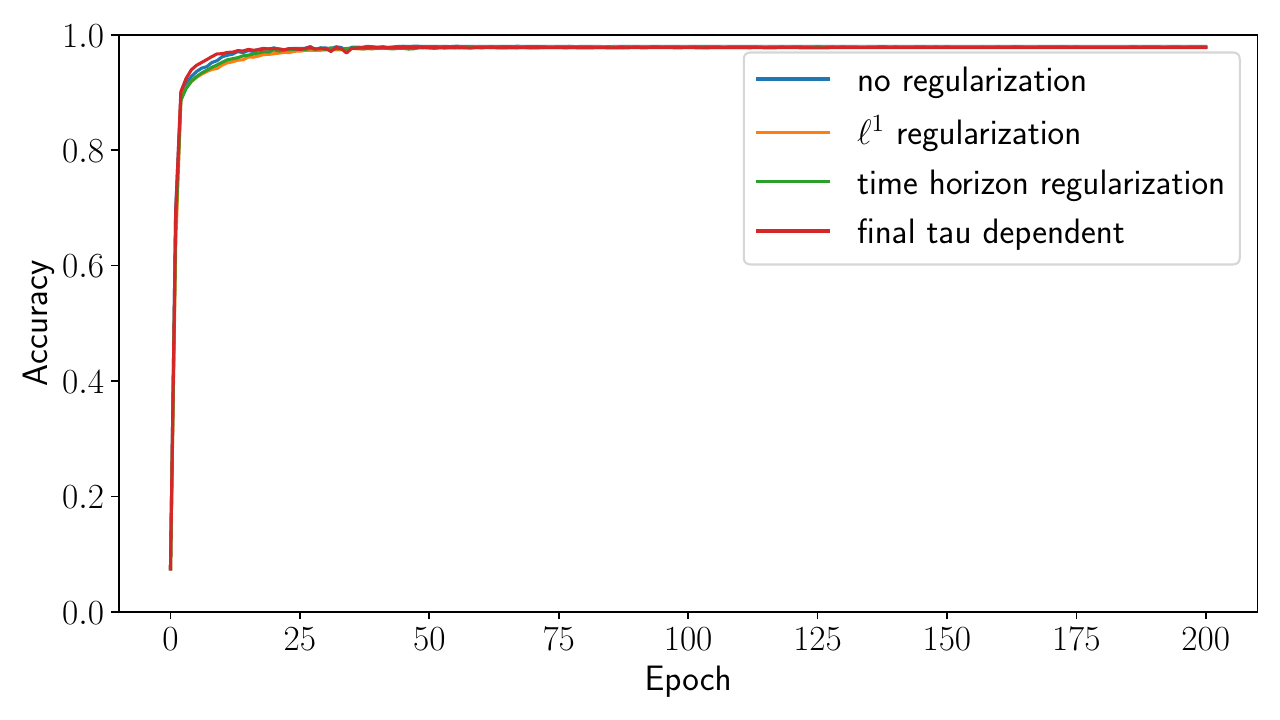}
	\includegraphics[width=0.495\textwidth]{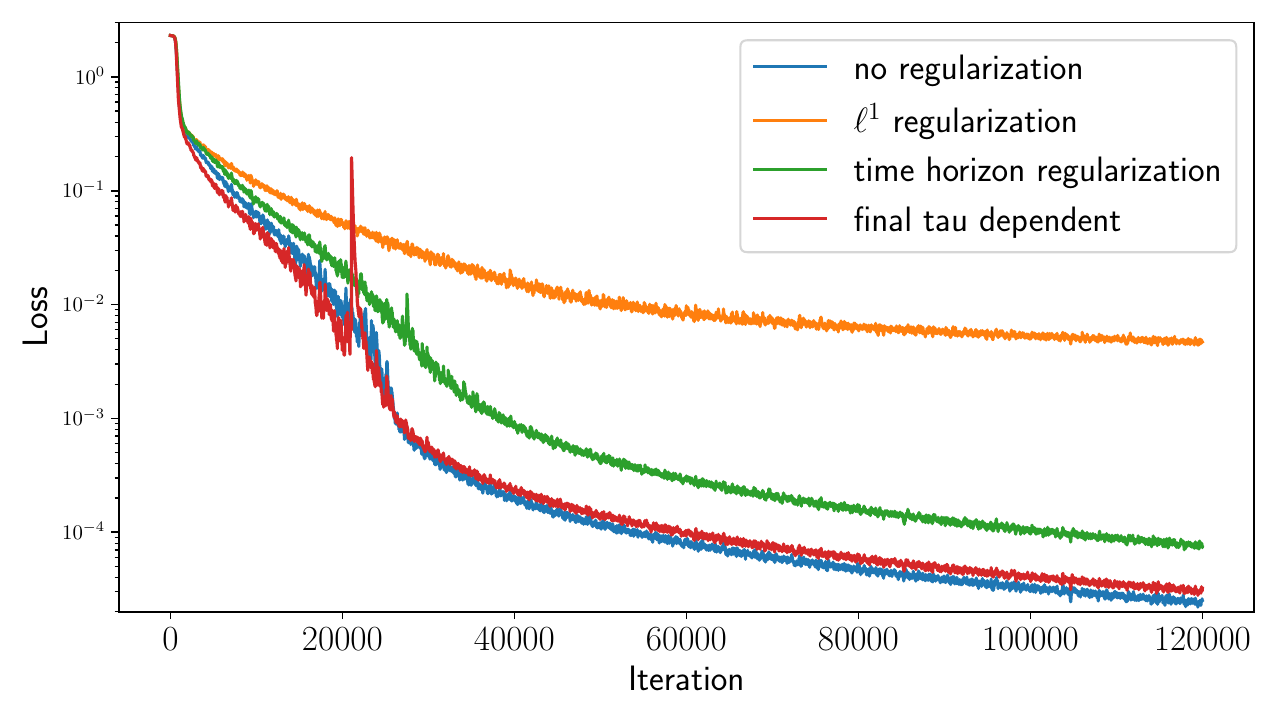}
		}
		\subfigure[Fashion MNIST]{\label{fig:a_t_r_2}
	\includegraphics[width=0.495\textwidth]{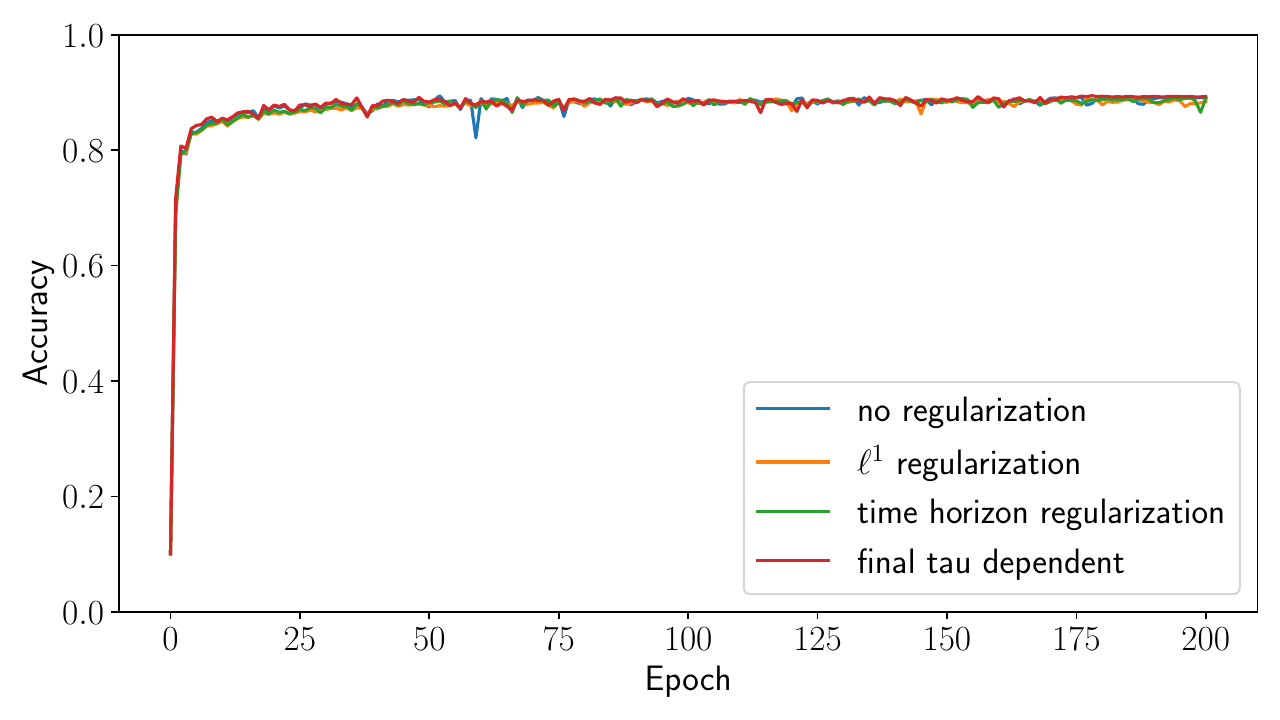}
	\includegraphics[width=0.495\textwidth]{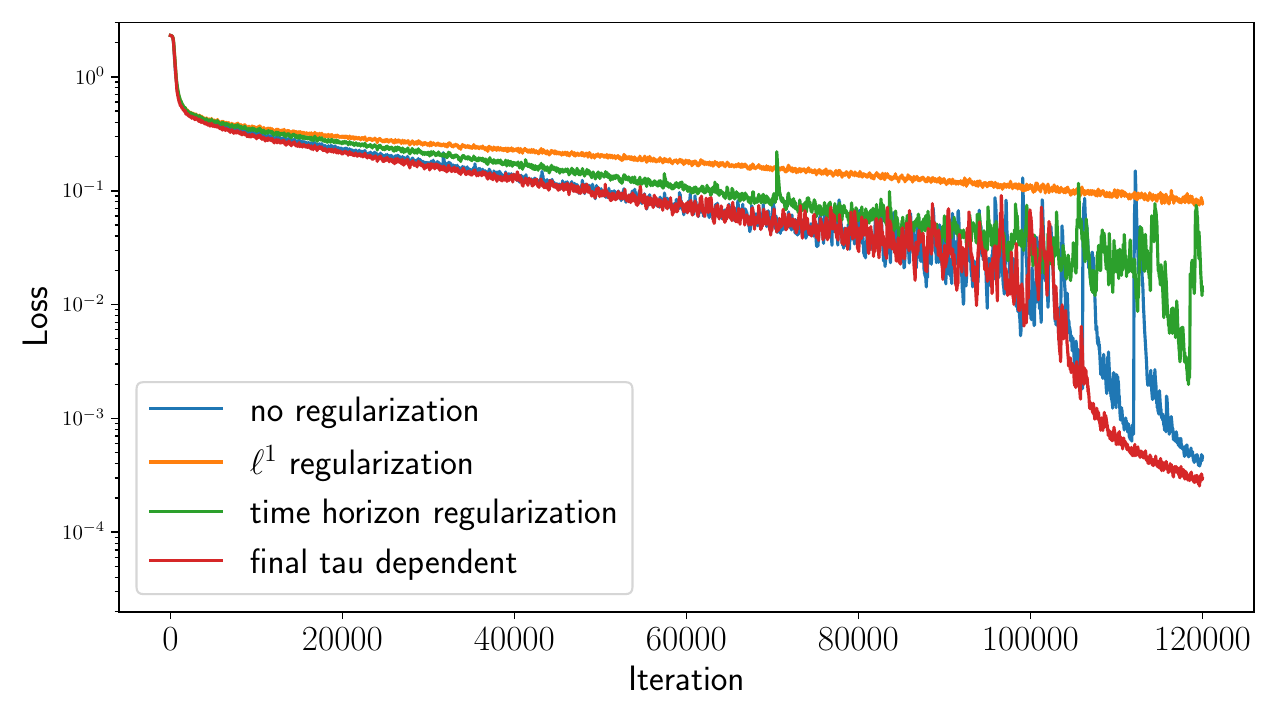}
		}
	\caption{Accuracies and cross entropy losses compared for different setups with trainable $\tau$ and ResNet architecture.}
	\label{fig:accuracies_trainable_resnet_MNIST}
\end{figure}
While the accuracies are similar for all four setups in time variable learning, we do observe different behaviors in the cross entropy loss decay during training, cf. \cref{fig:accuracies_trainable_resnet_MNIST} for the results when considering a ResNet architecture. 
However, recall that in the cases with $\ell^1$ regularization and time horizon regularization the loss function contains a regularization term, which explains that they deliver comparatively larger cross entropy loss values than in the cases without an added term. 
Furthermore, in \cref{fig:accuracies_trainable_fdnn} we have collected the observations with Fractional DNN architecture. 
Here, the setup with time horizon regularization admits a slightly lower accuracy during the first 100~epochs and a larger loss value throughout training. 
A potential explanation is that time horizon regularization in Fractional DNN is more restrictive than $\ell^1$ regularization and final $\tau$ dependency. 
In combination with the projection to ensure $\tau{\ell} > 0$ during training when Fractional DNN architecture is employed, it ensures that any $\tau{\ell} \ge T = 1$ will be penalized heavily, hence the values of the time step sizes remain smaller than in other setups, \cf \cref{fig:tau_behavior}. 
By comparing \cref{fig:accuracies} with \cref{fig:accuracies_trainable_resnet_MNIST,fig:accuracies_trainable_fdnn}, we see that even with regularization or an additional constraint on the time variables, training the chosen network architectures with variable~$\tau$ has advantages over training the same architectures with fixed~$\tau$. 

\begin{figure}[ht]
	\centering 
	\subfigure[no regularization]{\label{fig:w_p_1}\includegraphics[width=0.32\textwidth]{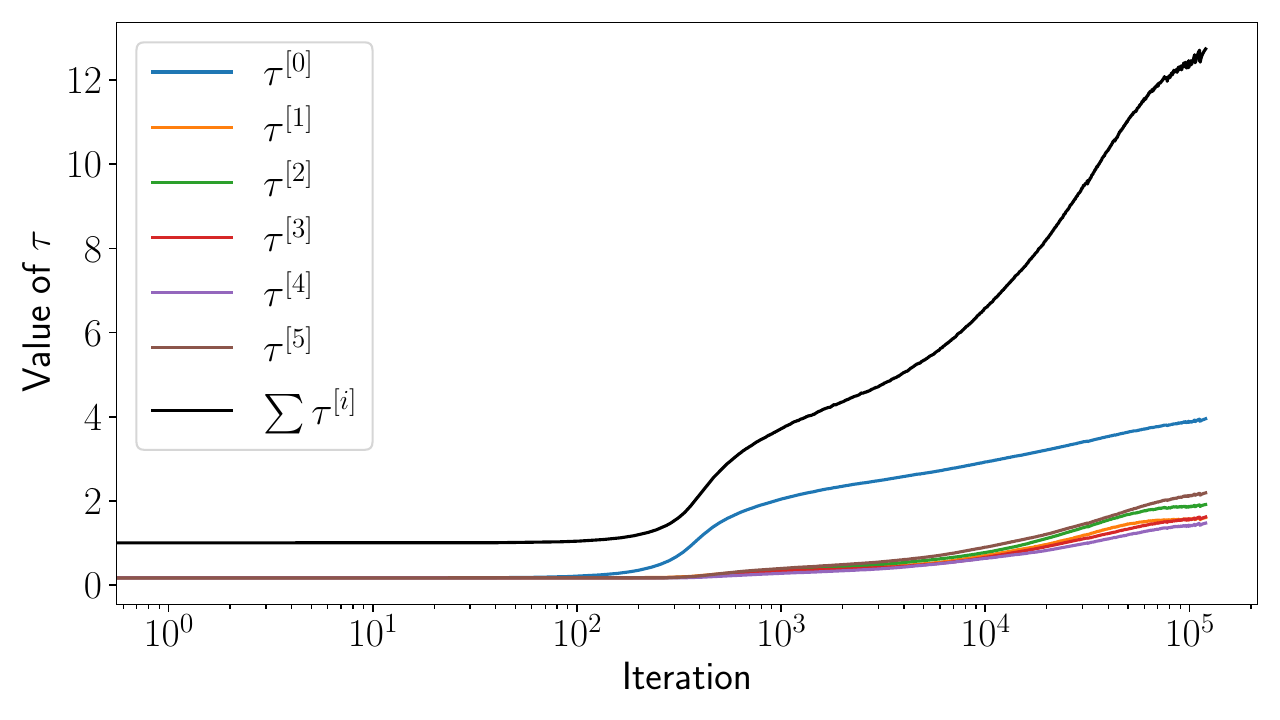}}
	\subfigure[$\ell^1$ regularization]{\label{fig:w_p_2}\includegraphics[width=0.32\textwidth]{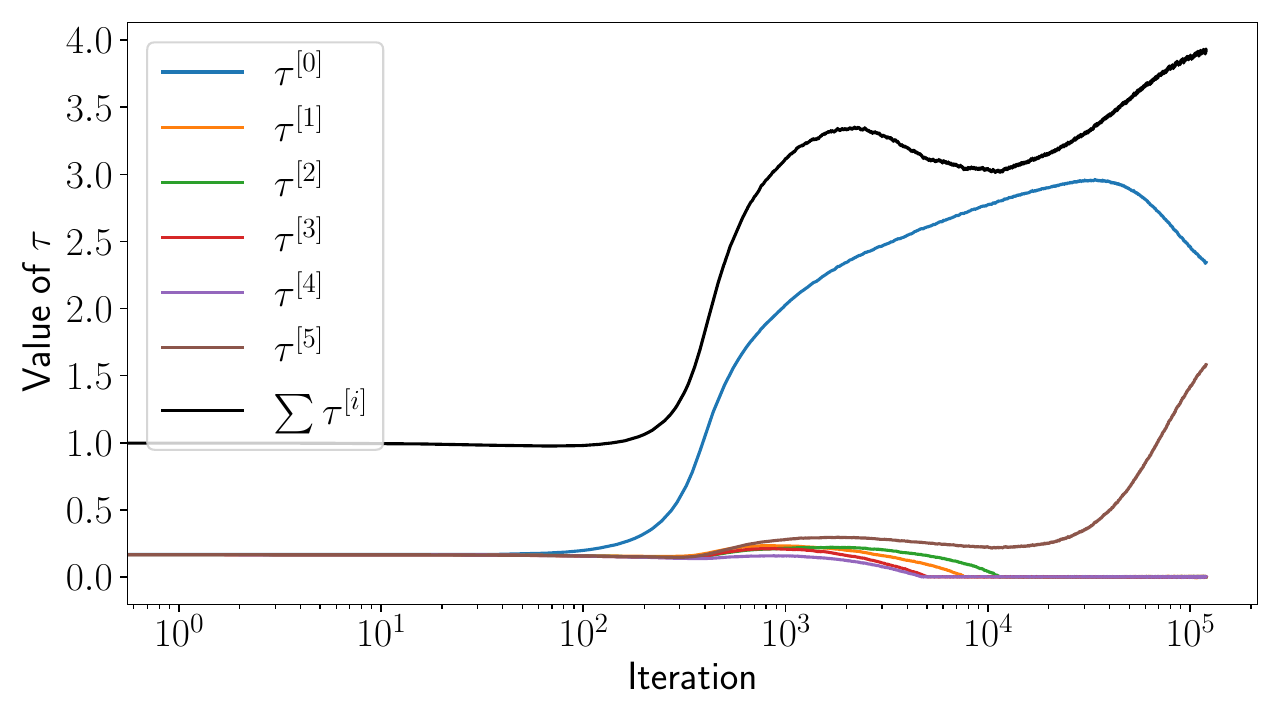}}
	\subfigure[time horizon regularization]{\label{fig:w_p_3}\includegraphics[width=0.32\textwidth]{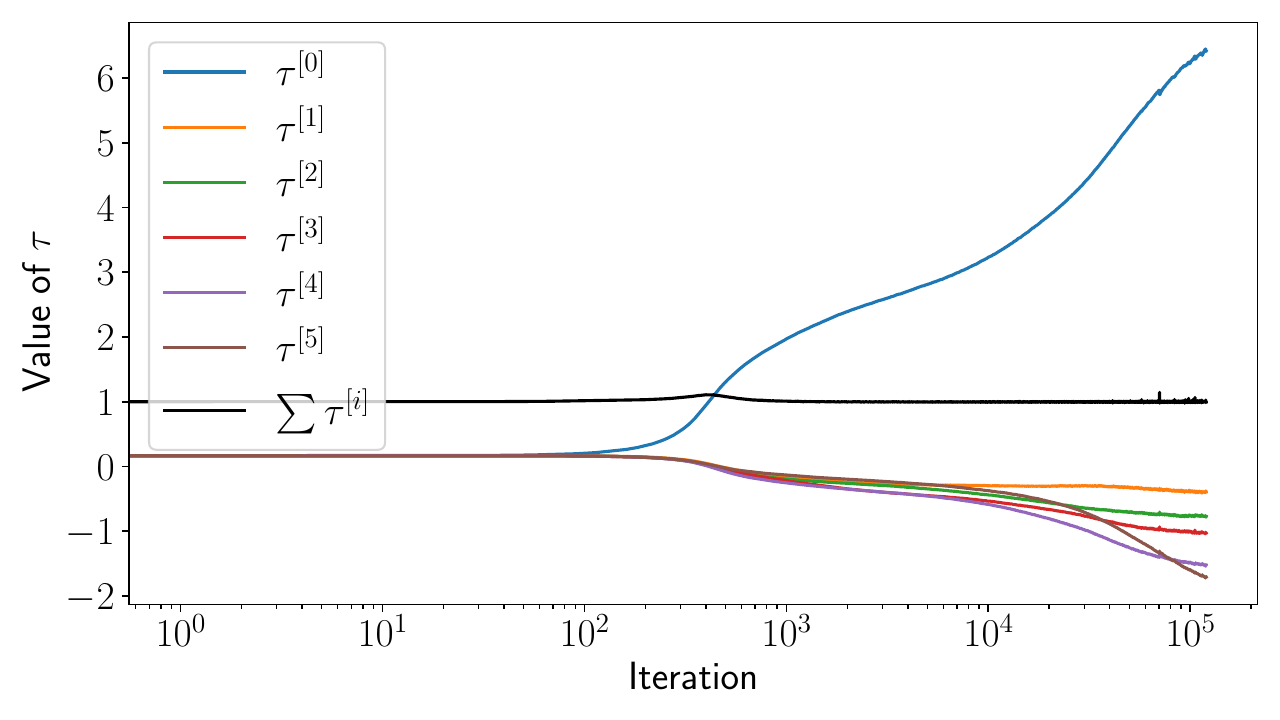}}
	\subfigure[final $\tau$ dependent]{\label{fig:w_p_4}\includegraphics[width=0.32\textwidth]{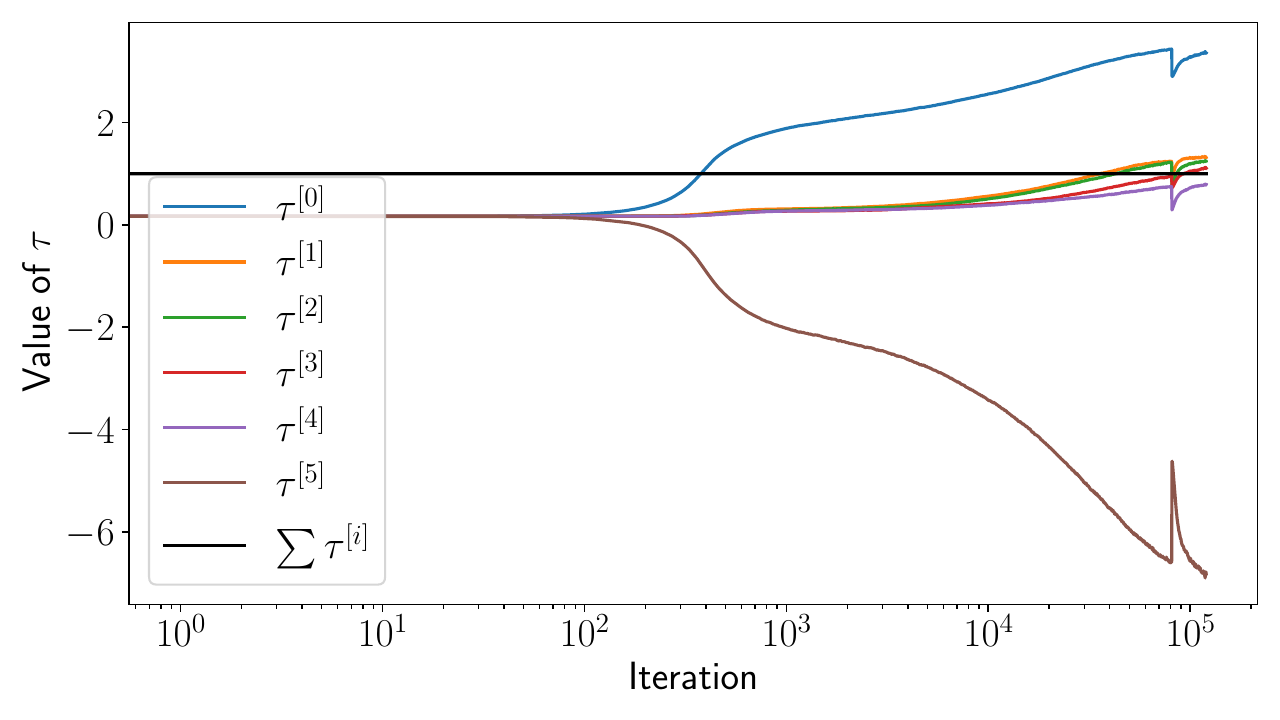}}
	\subfigure[$\ell^1$ and time horizon regularization]{\label{fig:w_p_5}\includegraphics[width=0.32\textwidth]{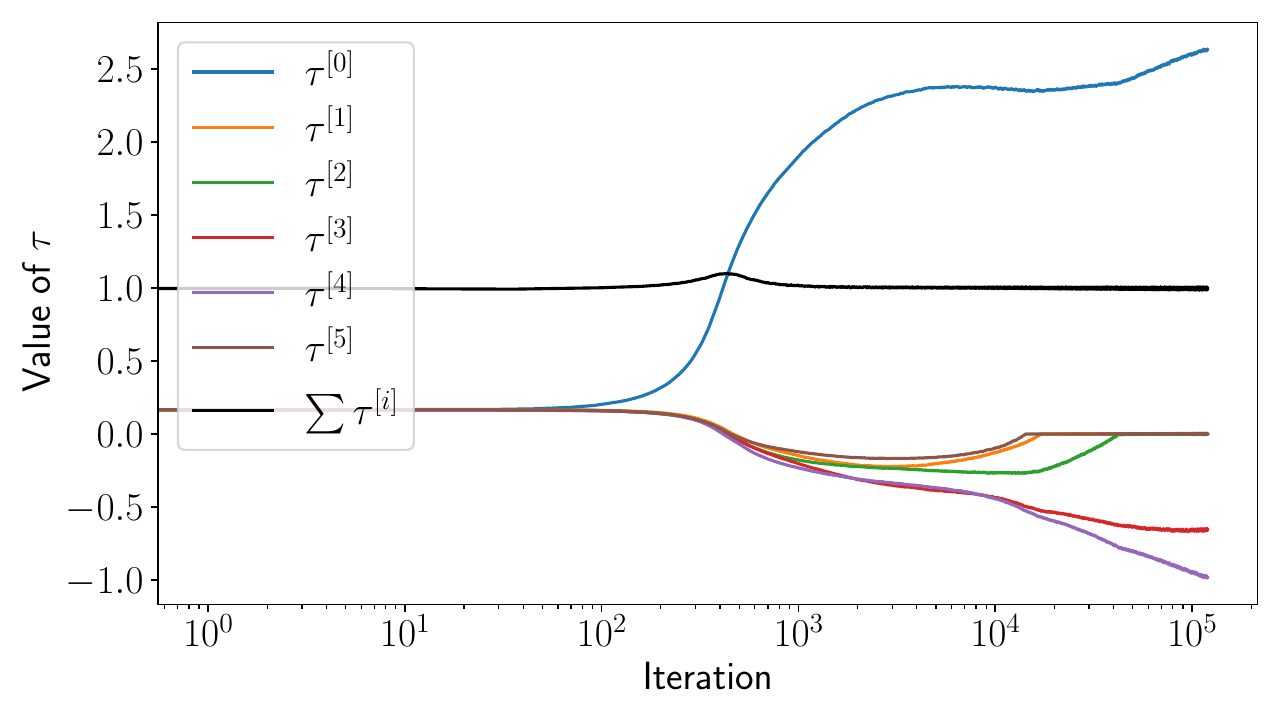}}
	\caption{Development of $\tau$ values over iterations for ResNet architecture with different regularization terms while learning to classify Fashion MNIST data. Please note that the scale on the vertical axis differs from plot to plot.}
	\label{fig:without_pruning}
\end{figure}

Next, we study the development of the time step size variables $\tau{0}, \ldots, \tau{5}$ during training. 
In \cref{fig:without_pruning} we display the development when training on the Fashion MNIST data set with ResNet architecture in different setups. 
In our example, we observe that $\tau{0}$ tends to be significantly larger than the other time step size variables in the network. 
Furthermore, the time horizon regularization ensures $\sum_{\ell=0}^{L-2} \tau{\ell} = T = 1$ in most iterations. 
In these setups, after a bit more than 100~iterations, we also see $\tau{\ell} < 0$ for $\ell \ge 1$.
Another interesting observation is that $\ell^1$ regularization drives some $\tau{\ell}$ to zero, where they remain. 
This clearly indicates that adaptive pruning can be beneficial, because the affected layers are redundant, \ie, $\layeroutput{\ell+1} = \layeroutput{\ell}$. 

In \cref{fig:tau_behavior} we display the development of the time step size variables when employing Fractional DNN architecture. 
Here, we have added a projection step into our algorithm to ensure $\tau{j} > 0$ and avoid the degeneracy of $a_{\ell,j}$. 
With final~$\tau$ dependency, we do not get $\tau{5} > 0$, since it is computed from the other time step size variables. 
Consequently, the variable~$\tau{5}$, which is initialized with a positive value, passes through zero. 
This can potentially blow up our experiment, however, we did not observe that. 
Without regularization, the values of all $\tau{\ell}$ are increasing with $\tau{0}$ being the largest. 
This behavior is hindered by $\ell^1$ and time horizon regularization. 
In comparison to the ResNet case, we do not observe $\tau{\ell}$ approaching zero when employing $\ell^1$ regularization. 
Additionally, due to the more complicated architecture, $\tau{\ell} \approx 0$ does not immediately indicate that a layer is redundant, hence we do not employ pruning with Fractional DNN architecture.
Furthermore, we observe that time horizon regularization in combination with the projection onto positive numbers clearly limits the development of $\tau{\ell}$. 
The sum exceeds $T=1$ at approximately iteration~\num{1000} and then goes back down while oscillating. 
Also, $\tau{1},\tau{2}, \tau{3}$ and $\tau{4}$ get close to zero, which can compromise computational stability. 
In conclusion, it seems that in cases with Fractional DNN architecture it may be more beneficial to abstain from regularization of the time step sizes. 

\begin{figure}[ht]
	\centering 
	\subfigure[$\ell^1$ regularization]{\label{fig:a_p_2}\includegraphics[width=0.32\textwidth]{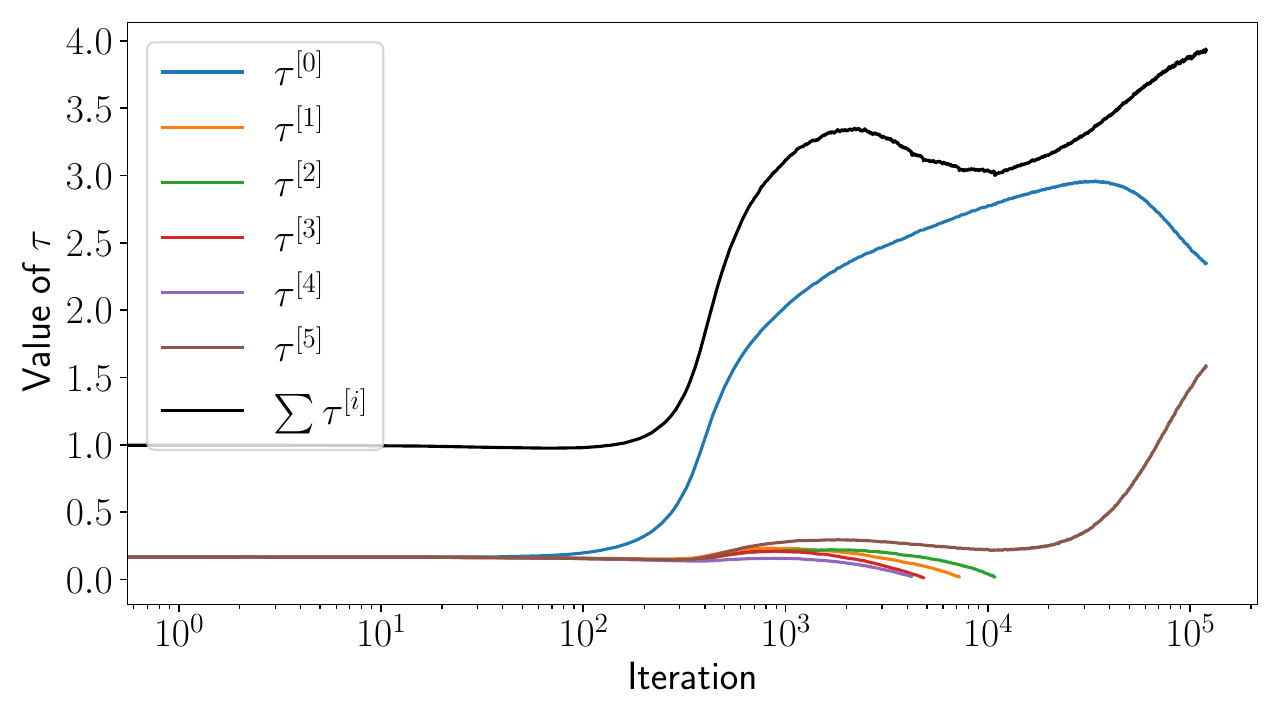}}
	\subfigure[time horizon regularization]{\label{fig:a_p_3}\includegraphics[width=0.32\textwidth]{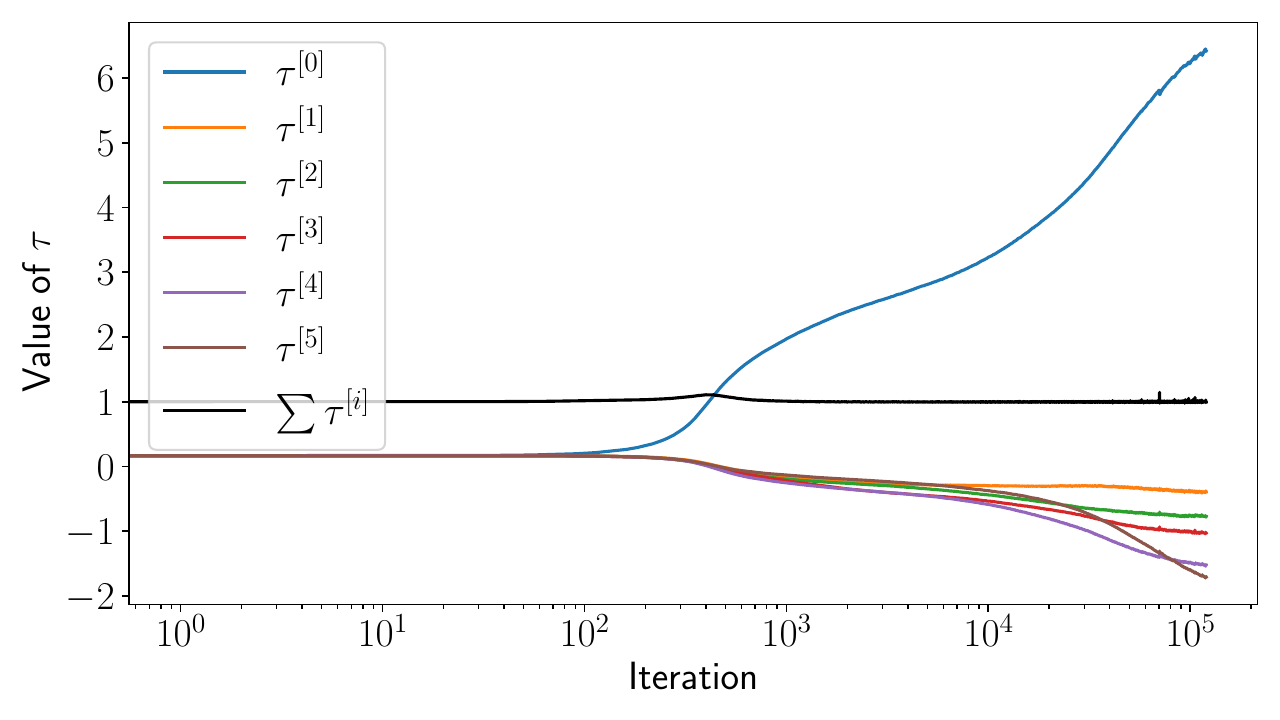}}
	\subfigure[$\ell^1$ and time horizon regularization]{\label{fig:a_p_5}\includegraphics[width=0.32\textwidth]{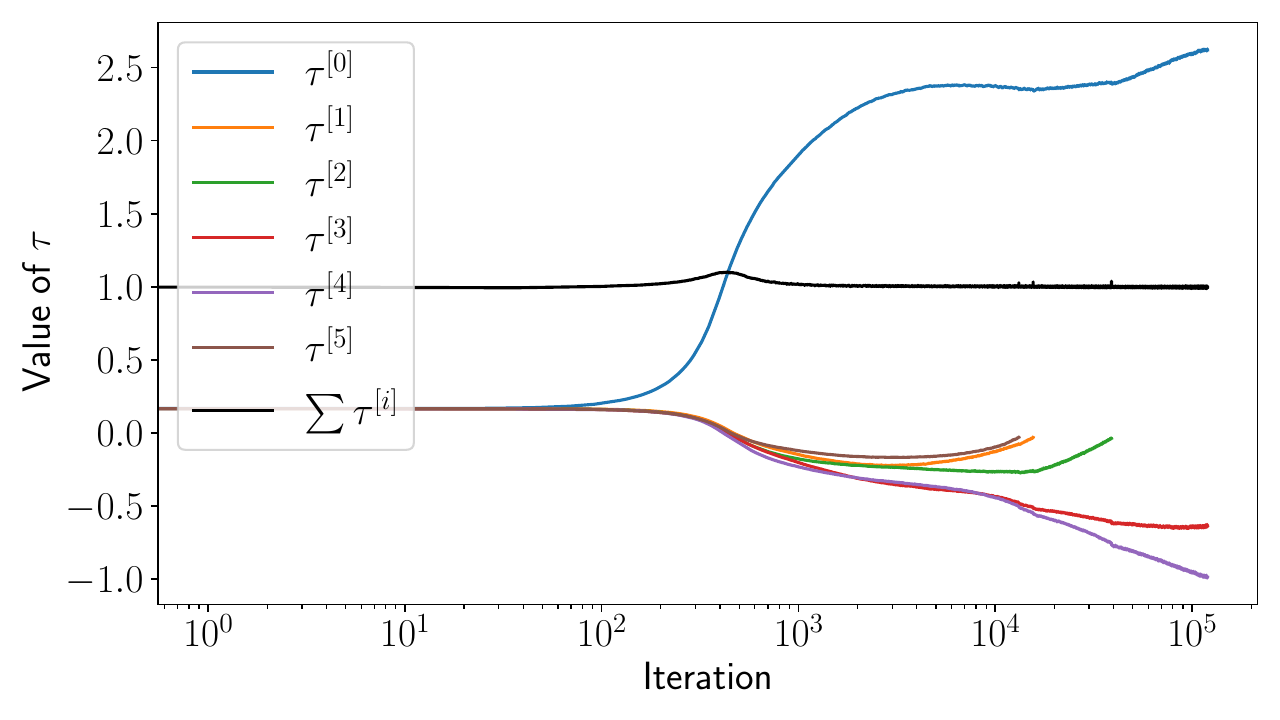}}
	\caption{Development of $\tau$ values over iterations with adaptive pruning ($\epsilon = 0.01$) for ResNet architecture with different regularization terms while learning to classify Fashion MNIST data. If a graph corresponding to a $\tau{\ell}$ terminates prematurely this indicates that pruning was executed, \eg, in (a) only $\tau{0}$ and $\tau{5}$ remain. Please note that the scale on the vertical axis differs from plot to plot.}
	\label{fig:adaptive_pruning}
\end{figure}

Finally, in \cref{fig:adaptive_pruning}, we display our results with adaptive pruning ($\epsilon = 0.01$) while training a ResNet to classify Fashion MNIST data. 
In cases with $\ell^1$ regularization this approach is successful and we can reduce our network size. 
In fact, every hidden layer omitted results in a reduction of the learning problem by \num{10101}~variables. 
The run time reduction is about half a minute when the total runtime without pruning was about $7$~minutes. 
This is expected to be more significant when considering larger architectures, where the network size has higher influence on the total runtime.

\appendix

\printbibliography

\end{document}